\documentclass[letterpaper, 10 pt, conference]{ieeeconf_RAS}  

\IEEEoverridecommandlockouts                              

\usepackage{amsmath} 
\usepackage{amssymb}  
\usepackage{booktabs}
\usepackage{color}
\usepackage{xcolor}

\usepackage[pdftex]{graphicx}
\usepackage{siunitx}
\usepackage{cite}
\usepackage{subcaption}
\usepackage[hyperfootnotes=false]{hyperref}
\usepackage{algorithm}
\usepackage{algpseudocode}
\usepackage{multirow}
\usepackage{verbatim}

\usepackage{xurl} 
\usepackage{glossaries}

\newacronym{HRI}{HRI}{human-robot interaction}
\newacronym{EBC}{EBC}{Expectation-Based Control}
\newacronym{MPC}{MPC}{Model Predictive Control}
\newacronym[longplural={degrees of freedom},shortplural={DoFs}]{DoF}{DoF}{degree of freedom}
\newacronym{SMU}{SMU}{Safe Motion Unit}
\newacronym{SSM}{SSM}{speed and separation monitoring}
\newacronym{PFL}{PFL}{power and force limiting}
\newacronym[longplural={inertial measurement units},shortplural={IMUs}]{IMU}{IMU}{inertial measurement unit}
\newacronym[longplural={evasive motions},shortplural={EMs}]{EM}{EM}{evasive motion}
\newacronym[longplural={Expectable Motion Units},shortplural={EMUs}]{EMU}{EMU}{Expectable Motion Unit}
\newacronym{FE}{FE}{Franka Emika}

\definecolor{lightblue_Matlab}{HTML}{00FFFF} 
\definecolor{blue_Matlab}{HTML}{0000FF}

\usepackage{stackengine}

\input{Settings/abkuerzung.sty}
\graphicspath{{graphics/}}
\DeclareGraphicsExtensions{.pdf,.jpeg,.png}

\renewcommand{\baselinestretch}{0.9}

\def\mytitle{Towards Connecting Control to Perception: High-Performance Whole-Body Collision Avoidance Using Control-Compatible Obstacles}
\title{\LARGE \bf \mytitle}

\def\myauthor{Moritz Eckhoff, Dennis Knobbe, Henning Zwirnmann, Abdalla Swikir, and Sami Haddadin}

\author{\myauthor
\thanks{
Funded by the Federal Ministry of Education and Research (BMBF) and the Free State of Bavaria under the Excellence Strategy of the Federal Government and the Länder. 
Funded by the German Research Foundation (DFG, Deutsche Forschungsgemeinschaft) as part of Germany’s Excellence Strategy – EXC 2050/1 – Project ID 390696704 – Cluster of Excellence “Centre for Tactile Internet with Human-in-the-Loop” (CeTI) of Technische Universität Dresden.
The authors acknowledge the financial support by the Bavarian State Ministry for Economic Affairs, Regional Development and Energy (StMWi) for the Lighthouse Initiative KI.FABRIK, (Phase 1: Infrastructure as well as the research and development program under grant no. DIK0249).
Please note that S. Haddadin has a potential conflict of interest as shareholder of Franka Emika GmbH.}
\thanks{All authors are with Technical University of Munich, Germany; TUM School of Computation, Information, and Technology, Department of Computer Engineering, Chair of Robotics and Systems Intelligence; 
Munich Institute of Robotics and Machine Intelligence (MIRMI).
AS and SH are also with Centre for Tactile Internet with Human-in-the-Loop (CeTI) of Technische Universität Dresden, Germany. AS is also with the Department of Electrical and Electronic Engineering, Omar Al-Mukhtar University, Albaida, Libya.
}
\thanks{Corresponding author: {\href{mailto:moritz.eckhoff@tum.de}{\tt\small moritz.eckhoff@tum.de}}}
}

\def\mysubject{}
\def\mykeywords{Whole-Body Motion Planning and Control; Force Control; Multi-Modal Perception for HRI}

\hypersetup{ 
pdftitle={\mytitle},
pdfsubject={\mysubject},
pdfauthor={\myauthor},
pdfkeywords={\mykeywords}
}

\usepackage{fancyhdr}
\fancypagestyle{firststyle}
{
   \fancyhf{}
   
\fancyhead[LO,LE]{\small{\textbf{2023 IEEE/RSJ International Conference on Intelligent Robots and Systems (IROS) [Preprint-Version]} \\\textbf{October 1-5, 2023. Detroit, USA}}}
\fancyfoot[LO,LE]{\copyright\footnotesize{\ 2023 IEEE. Personal use of this material is permitted. Permission from IEEE must be obtained for all other uses, in any current or future media, including reprinting/republishing this material for advertising or promotional purposes, creating new collective works, for resale or redistribution to servers or lists, or reuse of any copyrighted component of this work in other works. DOI: \href{https://doi.org/10.1109/IROS55552.2023.10342515}{10.1109/IROS55552.2023.10342515}}}
\fancyfoot[RO,RE]{\thepage}
}
\fancypagestyle{elsestyle}
{
\fancyhf{}

\fancyfoot[LO,LE]{\copyright\footnotesize{\ 2023 IEEE. Personal use of this material is permitted. Permission from IEEE must be obtained for all other uses, in any current or future media, including reprinting/republishing this material for advertising or promotional purposes, creating new collective works, for resale or redistribution to servers or lists, or reuse of any copyrighted component of this work in other works. DOI: \href{https://doi.org/10.1109/IROS55552.2023.10342515}{10.1109/IROS55552.2023.10342515}}}
\fancyfoot[RO,RE]{\thepage}
}

\usepackage{color}

\begin{document}

\maketitle
\thispagestyle{firststyle}
\pagestyle{elsestyle}

\begin{abstract}
One of the most important aspects of autonomous systems is safety. 
This includes ensuring safe human-robot and safe robot-environment interaction when autonomously performing complex tasks or in collaborative scenarios. 
Although several methods have been introduced to tackle this, most are unsuitable for real-time applications and require carefully hand-crafted obstacle descriptions.
In this work, we propose a method combining high-frequency and real-time self and environment collision avoidance of a robotic manipulator with low-frequency, multimodal, and high-resolution environmental perceptions accumulated in a digital twin system.
Our method is based on geometric primitives, so-called primitive skeletons. 
These, in turn, are information-compressed and real-time compatible digital representations of the robot's body and environment, automatically generated from ultra-realistic virtual replicas of the real world provided by the digital twin.
Our approach is a key enabler for closing the loop between environment perception and robot control by providing the millisecond real-time control stage with a current and accurate world description, empowering it to react to environmental changes.
We evaluate our whole-body collision avoidance on a 9-DOFs robot system through five experiments, demonstrating the functionality and efficiency of our framework.
\end{abstract}



\vspace{-0.15cm}
\section{INTRODUCTION}

\subsection{Motivation}
\begin{figure}[tpb]
    \centering
    \small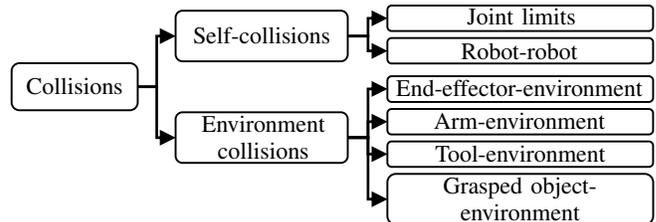
    \vspace{-0.5cm}
    \caption{Taxonomy of possible robot collisions.}
    \label{fig:collision_taxonomy}
    \vspace{-0.65cm}
\end{figure}

For a while now, the use of autonomous systems in close proximity to humans, for example, in service robotics or individual automation of laboratory processes, has been increasing. As a result, safety and robustness are becoming increasingly important for creating useful collaborative workspaces. Consequently, safe physical human-robot interaction (pHRI) must be ensured to enable complex robotic tasks in mostly unstructured environments. 
Crucial is collision avoidance, which can be implemented in offline motion planning or interactively in real-time robot control. 
Predictive collision-free motion planners allow for globally optimal paths, but constraints from confined workspaces and nonlinearities, e.g., robot kinematics or dynamics, lead to long computation times or infeasible optimization problems. Thus, simplifications, such as joint space-only or end-effector pose-only motion planning, are typically applied, often ignoring obstacles or the robot's posture.
Moreover, motion planners are not designed to react immediately and safely to unplannable events such as unavoidable human-robot collisions, after which resulting avoidable collisions must be prevented. However, this ability is essential in, e.g., laboratory environments, where hazardous goods can be spilled and overturned.
Hence, we advocate combining globally optimal motion planning with our reactive whole-body and environment-aware collision avoidance control to meet the real-time requirement for safe pHRI \cite{Kirschner2021} when performing robotic tasks.
In the past, full collision avoidance implementations in control, according to Fig. \ref{fig:collision_taxonomy}, were rare. This is due to the required distances to mostly unknown environmental objects with unknown poses and complex geometries.
Combined with the strict real-time constraints for safe pHRI and the possibility of environmental objects changing position, collision avoidance becomes challenging.
In this paper, we propose a solution to the problems of whole-body and environment-aware collision avoidance control and test the functionality with a 9-degrees of freedom (9-DOFs) mobile robot.

\subsection{Related Work}
\label{sec:RelatedWork}
\begin{table*}[tpb]
    \centering
    \caption{Collision avoidance pipelines with data processing\,(P), control\,(C), robot\,(R), sensor\,(S), and digital twin\,(T)}
    \vspace{-0.2cm}
    \begin{tabular}{lp{4.8cm}|p{2.3cm}p{2.7cm}p{2.65cm}p{2.55cm}}
        \toprule
        \multirow{10}{*}{\rotatebox[origin=c]{90}{Method}}
        & Name & Pipeline 1 &  Pipeline 2  & Pipeline 3 & Pipeline 4 \\
        & Collision avoidance technique & CBF &  APF  & APF & APF \\
        & Distance determination by & CBF &  Data Processing & Digital Twin & Control \\ 
        & Collision avoidance realized in & CBF &  Control & Digital Twin & Control \\ 
        & Equivalent control loop & \raisebox{-0.5\totalheight}{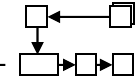} & \raisebox{-0.5\totalheight}{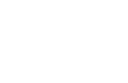}  & \raisebox{-0.5\totalheight}{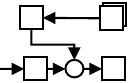} &  \raisebox{-0.5\totalheight}{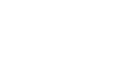} \\ 
        & Applied in & \cite{Chiu2022} &\cite{Scimmi2021} & Our alternative pipeline & Our proposed pipeline\\ 
        \midrule
        \multirow{7}{*}{\rotatebox[origin=c]{90}{Performance}}
        & Computational load of real-time control & None &  Low  & None & Medium \\ 
        & Delay between sensor and control & Low &  Low  & Low & Low \\ 
        & Risk of discontinuous distances & High &  High  & Low & Low \\ 
        & Risk of discontinuous torques & High &  High  & High & Low \\ 
        & Risk of unseen obstacles & Low &  High  & Low & Low \\ 
        & Collision avoidance in \SI{}{\milli\second} real-time &  No  & Not inherently provided  & If properly synced & Yes \\ 
        & Deploy intelligent planning decisions & Not possible & Not possible & Possible & Possible \\
        \bottomrule
    \end{tabular}
    \label{tab:Comparison}
    \vspace{-0.6cm}
\end{table*}

A recent approach for collision avoidance is using Control Barrier Functions (CBFs) \cite{Ames2019,Chiu2022}. Since they consider robot dynamics, the absence of collisions can be mathematically proven when applying CBFs. However, they rely on an optimization problem, which can conflict with the hard millisecond real-time constraint from robot control.
Alternatively, learning-based methods have been introduced in \cite{FigueroaFernandez2018,Koptev2021,Fang2015a}. However, they require retraining after environment changes, making them unsuitable for dynamic and unstructured workspaces.
Artificial Potential Fields (APFs) \cite{Khatib1985}, in turn, can immediately react to such changes. Since the algorithm includes no optimization problem, it is always convergent and fast to compute.
Although not formally provable, the robust APF algorithm has been refined into the skeleton algorithm \cite{DeSantis2007b,Dietrich2011}, which uses swept spheres \cite{Beaumont1989} to efficiently describe robots only by points and lines.
In contrast to swept spheres, the authors in \cite{Jang2021a} used offline point clouds to describe the robot. This method, however, produces discontinuities in the minimal distances between objects. Nevertheless, they were the first to utilize the additional DOFs from their non-holonomic mobile base to avoid self and environment collisions. 
All of the methods described above require 3D distances for collision detection. Sensor solutions have been presented in \cite{Fan2021,Scimmi2021,Safeea2019}. However, direct use of sensor data can lead to discontinuities due to the different frequencies of sensor and robot control as well as to unseen obstacles caused by shadowing and blind spots. 
Alternatively, \SI{1}{\kilo\hertz} real-time capable analytical vector geometry has been used in \cite{Jang2021a,Beaumont1989,DeSantis2007b,Dietrich2011}. 
A database or map storing all obstacle positions is required to apply this. 
Using a database enables distance calculations to currently unobserved obstacles and resolves the shadowing and blind spot issues direct sensor solutions have.
Databases can be generated by, e.g., autonomous mapping as in \cite{Marton2008}, where kitchen objects are fitted into a measured point cloud. 
Another possibility is to connect directly to the virtual environment representation of a so-called digital twin system \cite{Kritzinger2018}. Various implementations and definitions of these types of systems are already available \cite{Kritzinger2018, Beetz2015, Beetz2018, Kumpel2021}. In our work, however, a digital twin is a cloud-based service that processes multimodal sensory data in a perceptual layer, thereby building a vast knowledge base about the real environment while representing this data with dynamically changing, ultra-realistic virtual replicas.


In this work, we present an APF-based collision avoidance algorithm that we connect to a digital twin to utilize its current virtual environment representation. For this, we first enhance the virtual environment representation by our replica abstraction layer (RAL) to automatically translate the twin's complex and detailed meshes into a control-compat\-ible environment representation consisting of primitive skeletons (PSs).
The PSs hold all object positions, orientations, and shapes. They are transmitted to control along with task-specific interaction parameters like a maximum repelling force and whether to ignore specific distance pairs to allow intentional manipulation.
We use this information in our second step for \SI{1}{\kilo\hertz} real-time, whole-body, and environment-aware collision avoidance control.
Our concept is tested on a 9-DOFs Intelligent Robotic Lab Assistant \cite{Knobbe2022} consisting of a 2-DOFs gantry system and a 7-DOFs Franka Emika (FE) manipulator.
By a minor sacrifice of computation time, our proposed novel connection between twin and control enabled by the RAL provides better collision avoidance performance than previous work, even in cluttered environments.


\section{CONNECTING CONTROL TO PERCEPTION}
To achieve environment-aware control, collision information must be provided, as discussed in Section \ref{sec:RelatedWork}. We identified potential for improvement in this area and thus propose two novel collision avoidance pipelines, which we evaluate below by comparing them with two existing ones on a conceptual level. Table \ref{tab:Comparison} summarizes the results.

\subsection{Concept Evaluation}
In \emph{Pipeline 1}, a sensor provides data that is converted into an environment map by data processing \cite{Chiu2022}. A CBF then uses the map for collision-free path optimization. Their algorithm, however, only supports sphere-to-sphere distances and is not real-time capable. \emph{Pipeline 2} applies APFs that can be used in \SI{1}{\kilo\hertz} real-time control \cite{Dietrich2011}. Table \ref{tab:Comparison}, however, compares with the non-real-time APF implementation of \cite{Scimmi2021} since they consider the environment. The authors calculate collision distances based on visual data in data processing. These distances are provided at \SI{30}{\hertz} for collision avoidance control.
Both so far introduced methods result in discontinuous collision distances because of under-sampling caused either by the lack of real-time capability of CBFs or by the distance calculation on low sensor frequency outside real-time control. This, in turn, leads to discontinuous forces and torques even in static environments, as the true distances change continuously when the robot moves. 
The use of a map in \emph{Pipeline 1} reduces the risk of unseen obstacles, whereas obstacles in \emph{Pipeline 2} can not be considered when they are out of sensor range. 

In our work, we aim to combine the real-time capability of APFs with the advantage of lower risk of unseen obstacles when using maps. Therefore, we apply APFs and connect real-time control to the virtual environment representation of a digital twin in both proposed collision avoidance \emph{pipelines 3 and 4}.
In \emph{Pipeline 3}, distances and collision-avoiding torques $\boldsymbol{\tau}_{\text{rep}}$ are computed inside the twin and are subsequently added to those from real-time control. This decoupling frees the real-time control from any computational load caused by collision avoidance but leads to synchronization problems between the digital twin and control. When providing the twin with the \SI{1}{\kilo\hertz} robot state update, it can compute real-time distances, reducing the risk of discontinuities. However, the risk of discontinuous torques $\boldsymbol{\tau}$ remains high due to the problematic synchronization.
Therefore, \emph{Pipeline 4}, which we implemented, performs distance calculations and collision avoidance inside the real-time control, which finally decreases discontinuities in both distances and torques. This method, however, requires a translation of the virtual environment representation into a control-compat\-ible environment representation for which we propose the RAL.
Moreover, while the \emph{Pipeline 1 and 3} cause no additional computational load to real-time control and \emph{Pipeline 2} only adds APFs to control, \emph{Pipeline 4} causes the highest computational load to real-time control due to the additional distance calculations besides APFs. Nevertheless, this increases the computation time only slightly because distance calculations between PSs are highly efficient, and the RAL reduces the number of PSs by only providing the currently reachable ones.
Since all four pipelines have a sensor data processing step, their delays between sensor and control remain approximately the same.

Up to this point, our proposed approach could also be realized with an isolated solution in which a data processing algorithm creates and updates a map, from which the RAL generates PSs.
However, we suggest connecting control to the twin, as twins hold the most comprehensive knowledge about the robot, task, and environment. The context of digital twins is, thus, highly beneficial for decision-making. Our proposed connection empowers it to directly deploy lower-frequency decisions to control and, thus, the real world. Examples of such decisions include whether the mobile base of our 9-DOFs robot system should avoid collisions or maintain its position. Also, specific distance pairs can be ignored to keep the ability to manipulate objects.
After grasping objects or picking up tools, the RAL can link their PSs to the end-effector frame to avoid collisions between bodies attached to the end-effector and the robot itself or the environment. 
The connection can also be used the other way around to teach PSs and improve the twin's virtual environment representation.
All these features are arguments in favor of connecting control to perception through a digital twin, taking advantage of its comprehensive knowledge and computation power, thereby bringing intelligence into control. The following sections explain our technical realization of \emph{Pipeline 4}. 

\subsection{Control Loop}
\begin{figure}[tpb]
    \centering
    \small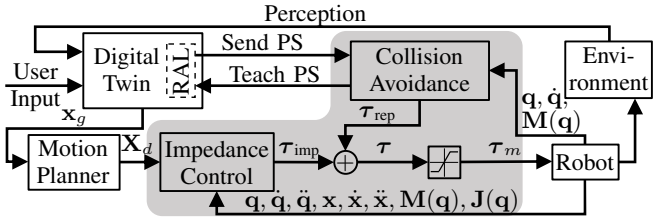
    \vspace{-0.4cm}
    \caption{Robot control loop connecting control to perception.}
    \label{fig:control_loop}
    \vspace{-0.6cm}
\end{figure}
Fig. \ref{fig:control_loop} shows our proposed control loop with its \SI{1}{\kilo\hertz} real-time control part highlighted in grey. The entry point into the loop is the digital twin, serving as a user interface where task sequences or even high-level goals can be defined. Based on this input and the perception of the environment, several models are built and updated in the twin. From these models, Cartesian goal poses $\mathbf{x}_g \in \mathbb{R}^{6}$ can be deducted, which are turned into desired trajectories $\mathbf{X}_d \in \mathbb{R}^{6\times h}$ with horizon $h$ by the motion planner. 
In our work, the twin model called virtual environment representation is of particular interest as our RAL generates environment PSs from it.
Those are exchanged through MongoDB with the collision avoidance module after environment changes. This connection is designed bidirectionally to also enable the teaching of PSs by guiding the robot's tool center point (TCP) to the corners of to-be-taught obstacles and determining their exact positions using the robot's forward kinematics.
In both modes, teaching and motion, the collision avoidance module gets real-time feedback, including the joint constellation and velocity $\mathbf{q}$ and $\dot{\mathbf{q}} \in \mathbb{R}^{N}$ and the mass matrix $\mathbf{M}(\mathbf{q})\in \mathbb{R}^{N\times N}$, where $N$ corresponds to the number of joints.
From this feedback, combined with the Denavit-Hartenberg (DH) parameters, all Cartesian link and joint poses are calculated and used either directly for teaching or in motion mode to determine collision distances.
The distances are needed to determine repelling torques $\boldsymbol{\tau}_{\text{rep}} \in \mathbb{R}^{N}$ based on our APF, which are then added to the impedance control torques $\boldsymbol{\tau}_{\text{imp}} \in \mathbb{R}^{N}$. The compliance of impedance control \cite{Siciliano.2010} allows path deviations when external forces from contacts are acting on the robot. As we aim to avoid collisions, we apply the external forces virtually before an actual contact occurs by adding $\boldsymbol{\tau}_{\text{rep}}$ to $\boldsymbol{\tau}_{\text{imp}}$. Their sums $\boldsymbol{\tau}$ are limited and finally sent to the robot as motor torques $\boldsymbol{\tau}_m$.
The second feedback loop in Fig. \ref{fig:control_loop} provides impedance control with the joint constellation $\mathbf{q}$, velocity $\dot{\mathbf{q}}$, and acceleration $\ddot{\mathbf{q}}$, the Cartesian end-effector pose $\mathbf{x}\in \mathbb{R}^{6}$, velocity $\dot{\mathbf{x}}$ ,and acceleration $\ddot{\mathbf{x}}$, the mass matrix $\mathbf{M}(\mathbf{q})$, and the Jacobian matrix $\mathbf{J}(\mathbf{q})\in \mathbb{R}^{6\times N}$, transforming joint constellations in end-effector poses.
Note that we use the classical model to describe the robot dynamics, which is
\begin{equation}
    \mathbf{M}(\mathbf{q})\ddot{\mathbf{q}} + \mathbf{C}(\mathbf{q},\dot{\mathbf{q}})\dot{\mathbf{q}} + \mathbf{g}(\mathbf{q}) =  \boldsymbol{\tau} + \boldsymbol{\tau}_{\text{ext}}\,,
\end{equation}
where $\mathbf{C}(\mathbf{q},\dot{\mathbf{q}})\in \mathbb{R}^{N\times N}$ is the Coriolis and centrifugal term, $\mathbf{g}(\mathbf{q})\in \mathbb{R}^{N}$ represents torques needed to compensate the gravity and $\boldsymbol{\tau}_{\text{ext}}\in \mathbb{R}^{N}$ corresponds to external torques.

Our implementation focuses on the RAL in the twin, the collision avoidance, and the connection between them.

\subsection{2D Primitive Skeletons for 3D Object Description}
\label{subsec:PrimitiveSkeleton}
To make oriented 3D replicas control-compatible, we first define oriented 2D PSs consisting of points
\begin{equation}
\label{eq:PrimitiveSkeletonSphere}
    \text{Point}: \,\Vec{\mathbf{x}} = \mathbf{O} \,,
\end{equation}
described by their origin vector $\mathbf{O} \in \mathbb{R}^{3}$, line segments
\begin{equation}
\label{eq:PrimitiveSkeletonCapsule}
    \text{Line}: \,\Vec{\mathbf{x}} = \mathbf{O} + s\mathbf{P}\,,
\end{equation}
described by their origin vector $\mathbf{O}$, direction vector $\mathbf{P} \in \mathbb{R}^{3}$, and the bounded scalar factor $s\in [0,1]$, and plane segments 
\begin{equation}
\label{eq:PrimitiveSkeletonPlane}
    \text{Plane}:\, \Vec{\mathbf{x}} = \mathbf{O} + s\mathbf{P} + t\mathbf{Q}\,,
\end{equation}
described by their origin vector $\mathbf{O}$, the perpendicular direction vectors $\mathbf{P}$ and $\mathbf{Q} \in \mathbb{R}^{3}$, and the bounded scalar factors $s, t\in [0,1]$. 
All scaling factors are bounded to keep their geometric bodies finite. 
\begin{figure}[tpb]
    \centering
    \begin{subfigure}{0.25\textwidth}
        \centering
        \small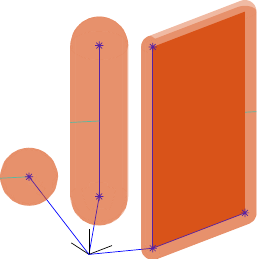
        \subcaption{The used primitives.}
        \label{subfig:object_occupatioon:3dbodies}
    \end{subfigure}
    \begin{subfigure}{0.22\textwidth}
        \centering
        \includegraphics[height = 44mm]{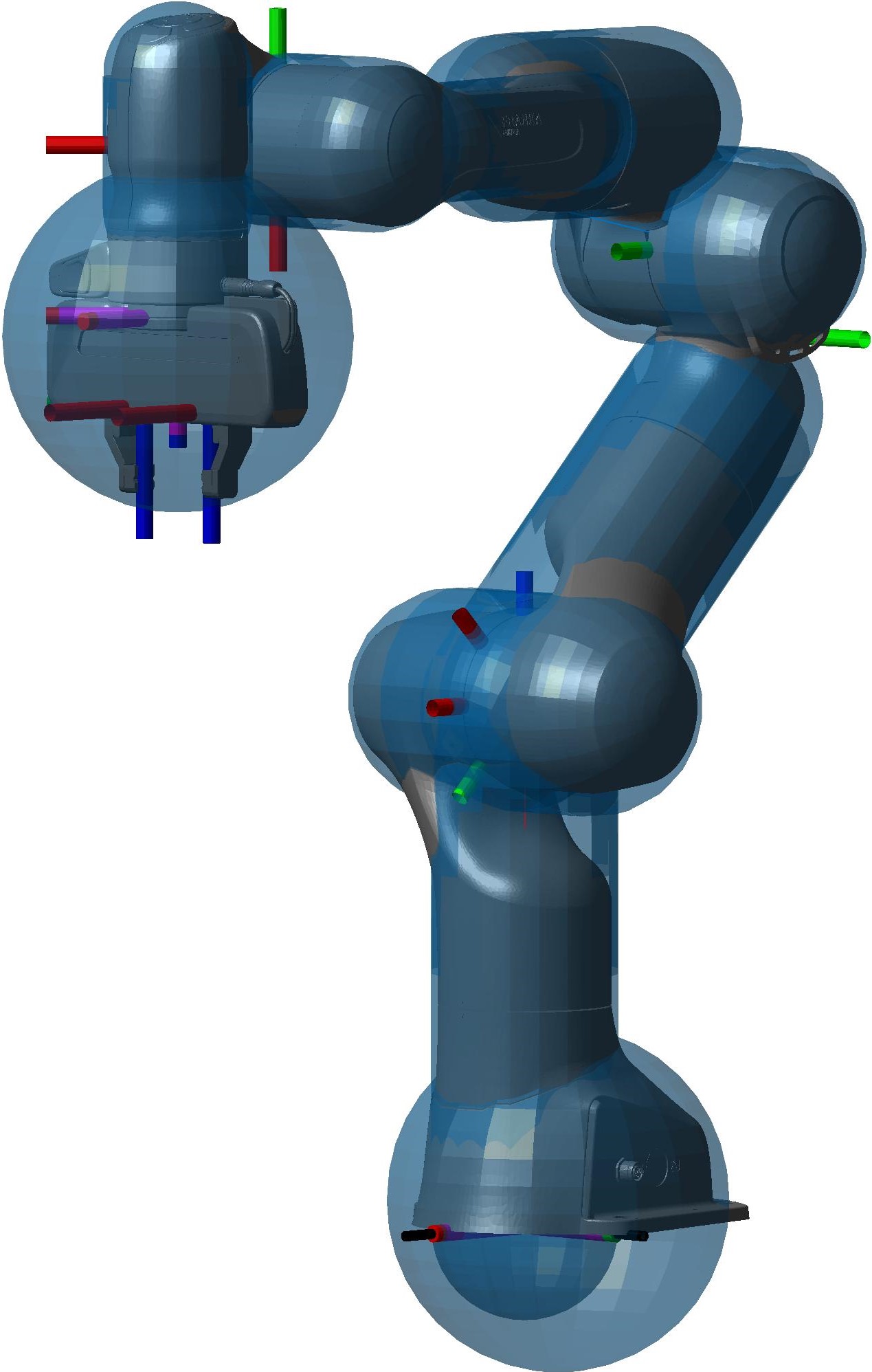}
        \subcaption{The FE robot.}
        \label{subfig:object_occupatioon:robot}
    \end{subfigure}
    \vspace{-0.1cm}
    \caption{(a) shows the 2D PSs in blue and their corresponding 3D primitive sphere, capsule and rounded cuboid in orange. (b) illustrates a tight 3D description of the FE robot.}
    \label{fig:object_occupatioon}
    \vspace{-0.6cm}
\end{figure}
By giving each point, line, and plane an individual radius $r$, the 2D PSs become control-compat\-ible 3D spheres, capsules, and rounded cuboids, as Fig. \ref{subfig:object_occupatioon:3dbodies} illustrates. The rounded surfaces support the generation of evasive motions, and the majority of objects the robot interacts with are well described with these three primitives. 
For the few others, triangles, i.e., rounded prisms, can be used, defined by plane skeletons with $s\in [0,1]$ and $t = f(s) = a s$. An application of the primitives is shown in Fig. \ref{subfig:object_occupatioon:robot}.

\subsection{The Twin's Replica Abstraction Layer (RAL)}
\begin{figure}[tpb]
    \centering
    \begin{subfigure}{0.49\columnwidth}
        \centering
        \includegraphics[width =\columnwidth]{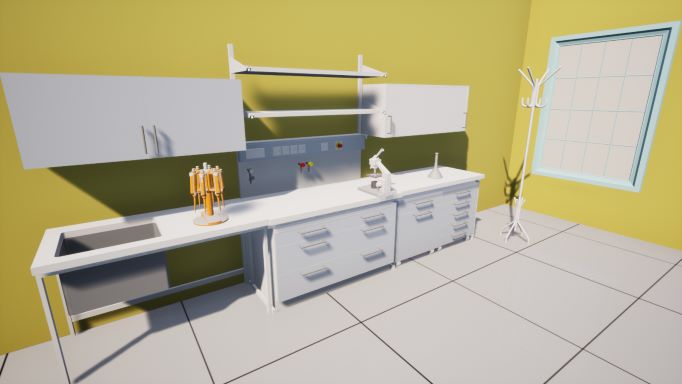}
        \subcaption{Mesh representation}
        \label{subfig:Twin:VirtEnvRep}
    \end{subfigure}
    \begin{subfigure}{0.49\columnwidth}
        \centering
        \includegraphics[width = \columnwidth]{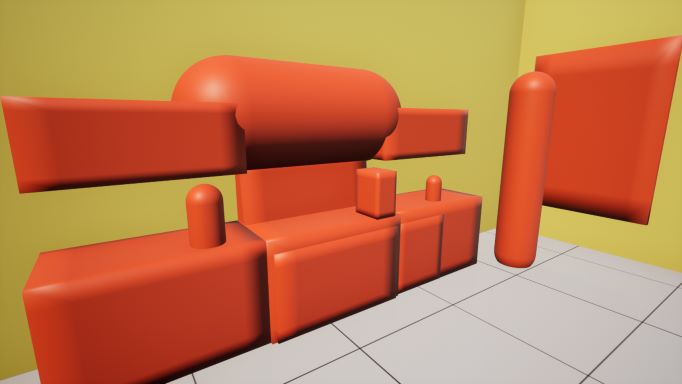}
        \subcaption{Primitive representation}
        \label{subfig:Twin:primEnvRep}
    \end{subfigure}
    \caption{(a) shows the virtual environment representation of a digital twin and (b) illustrates how the same environment is described using primitives generated by the RAL.}
    \label{fig:Twin}
    \vspace{-0.7cm}
\end{figure}

In this section, we enhance the RAL to the virtual environment representations of digital twins, illustrated in Fig. \ref{subfig:Twin:VirtEnvRep}. The RAL auto-generates a control-compat\-ible environment representation, shown in Fig. \ref{subfig:Twin:primEnvRep}, to enable high-performance whole-body collision avoidance.
To do so, we apply two simplifications to our RAL algorithm.
First, we assume that each mesh in our virtual environment representation is well approximated by only one primitive. 
This is important as one challenge in automatic shape approximation is the pre-approximation segmentation \cite{Bae2012}, defining how often and where a mesh needs to be divided into sub-meshes for an optimal approximation result. In other words, segmentation defines the optimal number of approximation primitives for each mesh and what part, i.e., sub-mesh, each primitive covers. For our collision avoidance application, we found that all small obstacles are well described by one primitive and thus do not require any segmentation. Only the approximation of large obstacles such as tables or laboratory supply islands can be significantly improved by segmentation. Since such obstacles, however, do not usually enter a workspace, they are known in advance. Users can correctly segment them beforehand if an advanced approximation is required.
Our second simplification is to exclude the robot from the shape approximation algorithm in the RAL, which is reasonable since the PSs of the robot links remain constant when described in their corresponding joint frame. This simplification even improves system performance as we embed the constant PSs in the collision avoidance module and update their world frame poses based on $\mathbf{q}$ with the frequency of \SI{1}{\kilo\hertz}.

Given these two simplifications, we implement the virtual environment representation shown in Fig. \ref{subfig:Twin:VirtEnvRep} in Unreal Engine~5, a state-of-the-art photorealistic gaming engine with C++ API. Unreal offers a built-in shape approximation module\footnote{\url{https://docs.unrealengine.com/5.0/en-US/API/Plugins/DynamicMesh/ShapeApproximation/} last accessed on 17.07.2023}, capable of approximating a mesh with one predefined primitive shape. Using this module, we fit four primitives to each currently reachable mesh. The approximation primitives are an aligned box (abox) where the box edges are parallel to the world coordinate axis, an oriented box (obox) not necessarily aligned to the world coordinate system, an oriented capsule (cap), and a sphere (sph). 
Although the aligned box approximation will always yield a solution that is at best equal to, but usually wider and thus worse than, that of the oriented box, we include it in the RAL. This is because its approximation algorithm is convergent, while those from the oriented box and capsule do not always converge. 
Since two of four approximations can fail, we detect failures by checking the volume $V_{p_k}$ of each primitive $p_k$ with $k \in \{\text{abox, obox, cap, sph}\}$ and set it to 
\begin{equation}
    V_{p_k} =
    \begin{cases} 
        V_{p_k}, & \text{if }V_{i} \leq V_{p_k}, \\
        \infty,  &\text{otherwise},
    \end{cases} \,
\end{equation}
where $V_{i}$ corresponds to the volume of the $i^{\text{th}}$ reachable mesh, determined according to \cite{ChaZhang2001}. The tightest valid approximation is the $p_k$ with minimal $V_{p_k}$. But to be able to give preference to specific, e.g., simpler primitives, we derive the optimal primitive approximation $p^{*}_i$ of mesh $i$ by solving the minimization problem 
\begin{equation}
    p^{*}_i = \underset{p_k}{\arg\min}{(V_{p_k}-V_{i})c(p_k)}\, ,
\end{equation}
with $c(p_k)$ providing the cost term associated with this primitive.
Finally, $p^{*}_i$ is decomposed into a PS. While this is straight forward for spheres and capsules, we represent boxes with $\|\mathbf{P}_i\|> \SI{5}{\centi\meter} \land \|\mathbf{Q}_i\|> \SI{5}{\centi\meter}$ by six planes having a radius of \SI{0}{\centi\meter} and those with $\|\mathbf{P}_i\|\leq \SI{5}{\centi\meter} \lor \|\mathbf{Q}_i\|\leq \SI{5}{\centi\meter}$ by only one plane lying in the middle of the shorter direction vector having a radius of half of the short direction vector's length. The visualization of the output of the RAL algorithm is shown in Fig. \ref{subfig:Twin:primEnvRep}.

\subsection{Communication between Twin and Control}
Reading and writing in a MongoDB enables the communication between RAL and control. 
The data exchange in JSON format happens whenever a motion starts or an environment PSs changes its pose in the RAL. It contains general information on whether the base is to be considered fixed or not, the total number of environment PSs, and the collision avoidance design values $F_{\text{max}}$, $d_\text{th}$, and $\zeta$. Moreover, it includes an array containing the environment PSs. Each entry again contains the PS name, its type (point, line, plane), its radius, the frame to which it is linked, a list with names of other PSs which this one ignores in collision avoidance, and the three vectors $\mathbf{O}_i$, $\mathbf{P}_i$, and $\mathbf{Q}_i$.
Finally, this communication enables real-time collision avoidance control to consider the environment PSs provided by the RAL and empowers the twin to deploy decisions directly into the control.
Having introduced this connection, one can easily extend the information exchange for future applications to put even more intelligent decisions into action.

\section{REAL-TIME COLLISION AVOIDANCE CONTROL}
\label{sec:CollisionAvoidanceControl}
Before each motion, the RAL provides the reachable environment PSs. Under their consideration, all relevant distance pairs $\vec{\mathbf{y}}_{i,j}$ are deducted. Afterwards, the following five steps are performed in each \SI{1}{\kilo\hertz} control loop iteration.
\paragraph{State Update}
Since the robot manipulator and the mobile base are movable, their PSs and those of the tool and the grasped object change position and orientation during motion. Therefore, they are defined in their $i^{\text{th}}$ robot frame and require an online transformation into the world frame $W$, using
\begin{equation}
\label{eq:GeneralTransformationEq}
    {}^W\mathbf{p}_i = {}^W\mathbf{t}_i(\mathbf{q})+{}^W\mathbf{R}_i(\mathbf{q}) {}^i\mathbf{p}_i \, ,
\end{equation}
where ${}^W\mathbf{R}_i(\mathbf{q}) \in \mathbb{R}^{3\times3}$ is the rotation matrix and ${}^W\mathbf{t}_i(\mathbf{q}) \in \mathbb{R}^{3}$ is the translation vector transforming a vector ${}^i\mathbf{p}_i$ from the $i^{\text{th}}$ frame to the world frame $W$. Applying \eqref{eq:GeneralTransformationEq} to the PS descriptions \eqref{eq:PrimitiveSkeletonSphere}-\eqref{eq:PrimitiveSkeletonPlane} results in the concrete state update
\begin{IEEEeqnarray}{rcl}
    \label{eq:StateUpdate1}
    {}^W\mathbf{O}_i & = & {}^W\mathbf{t}_i(\mathbf{q}) + {}^W\mathbf{R}_i(\mathbf{q}){}^i\mathbf{O}_i\, ,\\
    \label{eq:StateUpdate2}
    {}^W\mathbf{P}_i & = & {}^W\mathbf{R}_i(\mathbf{q}) {}^i\mathbf{P}_i\, ,\\
    \label{eq:StateUpdate3}
    {}^W\mathbf{Q}_i & = & {}^W\mathbf{R}_i(\mathbf{q}) {}^i\mathbf{Q}_i\, .
\end{IEEEeqnarray}
\paragraph{Distance Calculation}
We describe distances $\vec{\mathbf{y}}_{i,j} \in \mathbb{R}^3$ as line segments in space 
\begin{equation}
\label{eq:GeneralDistance}
    \text{Distance}: \,\Vec{\mathbf{y}}_{i,j} = \mathbf{U}_{i,j} +v_{i,j}\mathbf{W}_{i,j} \, , 
\end{equation}
with an origin vector $\mathbf{U}_{i,j} \in \mathbb{R}^3$ on PS $i$, a direction vector $\mathbf{W}_{i,j} \in \mathbb{R}^{3}$, pointing from $\mathbf{U}_{i,j}$ onto the closest point of PS $j$, and an associated scaling factor $v_{i,j} \in [0,1]$. They are calculated using analytical vector geometry. For all combinations of distance pairs, we start with the difference between PS $i$ and $j$ to obtain the direction vector
\begin{equation}
\label{eq:distace_direction}
    \mathbf{W}_{i,j} = \Vec{\mathbf{x}}_j - \Vec{\mathbf{x}}_i \, .
\end{equation}
This equation is directly applicable to point-point distances, while for lines or planes, the scaling factor values for minimal distance need to be determined first. This is done using the perpendicularity condition 
\begin{equation}
\label{eq:PerpendicularCondition}
    \mathbf{W}_{i,j} \perp \Vec{\mathbf{x}}_i \land \mathbf{W}_{i,j} \perp \Vec{\mathbf{x}}_j \, ,
\end{equation}
which holds true if we temporarily assume the PSs $i$ and $j$ to have an infinite length as the minimal distance vector, then is always perpendicular to both PSs. Note this assumption of cause requires a subsequent bounding or limitation step, which we will describe later.
Knowing that the dot product of two perpendicular vectors gives zero, we can derive different systems of equations from \eqref{eq:PerpendicularCondition} depending on which distance pair (line-line, point-plane, line-plane, etc.) we have.
For point-line distances, condition \eqref{eq:PerpendicularCondition} results in 
\begin{equation}
\label{eq:SystemOfEquationsPointLine}
    (\Vec{\mathbf{x}}_j - \Vec{\mathbf{x}}_i) \cdot \mathbf{P}_j = 0 \,\,\text{or}\,\, (\Vec{\mathbf{x}}_j - \Vec{\mathbf{x}}_i) \cdot \mathbf{P}_i = 0 \, ,
\end{equation}
for line-line distances, using \eqref{eq:PerpendicularCondition} we have 
\begin{equation}
\begin{aligned}
\label{eq:SystemOfEquationsLineLine}
    (\Vec{\mathbf{x}}_j - \Vec{\mathbf{x}}_i) \cdot \mathbf{P}_i = 0 \, ,\\
    (\Vec{\mathbf{x}}_j - \Vec{\mathbf{x}}_i) \cdot \mathbf{P}_j = 0\, ,
\end{aligned}
\end{equation}
and for point-plane distances, based on \eqref{eq:PerpendicularCondition} one gets 
\begin{equation}
\begin{aligned}
\label{eq:SystemOfEquationsPointPlane}
    (\Vec{\mathbf{x}}_j - \Vec{\mathbf{x}}_i) \cdot \mathbf{P}_j = 0 \, ,\\
    (\Vec{\mathbf{x}}_j - \Vec{\mathbf{x}}_i) \cdot \mathbf{Q}_j = 0\, .
\end{aligned}
\end{equation}
By solving \eqref{eq:SystemOfEquationsPointLine}-\eqref{eq:SystemOfEquationsPointPlane} for $s_i$, $s_j$, and $t_j$, their minimal distance values are derived, which can lie outside of their defined domain $[0,1]$ as we until now assumed the PSs to be infinite. 
To obtain the scaling factor values for the minimal distances between the finite PSs, we first set only one scaling factor to its closer domain limit if the current value is outside its limits. If the collision pair has a second factor, we fix the first factor, thus simplifying to a point-line problem according to \eqref{eq:SystemOfEquationsPointLine}, and then recalculate the second factor. Finally, the second factor is also limited.
The line-plane distance is the last required distance pair we have not discussed so far. Since infinite lines and planes can only be parallel or intersecting, their distance is either always constant and has no minimum, or the minimal distance point is that of the intersection. In the case of parallelism, the center of the overlapping sections of the PSs is used as a minimal distance point, while in the case of intersection, the intersecting point $\mathbf{p}_{\text{int}}$ is determined by solving
\begin{equation}
\label{eq:SystemOfEquationsLinePlane}
\mathbf{O}_i + s_i\mathbf{P}_i = \mathbf{O}_j + s_j\mathbf{P}_j + t_j\mathbf{Q}_j \, ,
\end{equation}
for $s_i=s_{\text{int}}$ and applying $\mathbf{p}_{\text{int}} = \mathbf{O}_i + s_{\text{int}}\mathbf{P}_i$. Based on case distinctions on $\mathbf{p}_{\text{int}}$, the line-plane distance problem can then be simplified to either a line-line or point-plane distance problem, which can easily be solved as already described.
Inserting the values of the minimal distance scaling factors of PS $i$ into \eqref{eq:PrimitiveSkeletonSphere}-\eqref{eq:PrimitiveSkeletonPlane} yields the vector $\mathbf{U}_{i,j}$, while $\mathbf{W}_{i,j}$ is calculated by inserting the values of PSs $i$ and $j$ into \eqref{eq:distace_direction}.
In order to obtain the later required 3D surface-to-surface distance $d_{i,j}$ from it we apply
\begin{equation}
    d_{i,j} = \|\mathbf{W}_{i,j}\| - r_i - r_j \, .
\end{equation}
The very last distance to be defined is the one between $\mathbf{q}$ and the joint limits, which we calculate as
\begin{equation}
\mathbf{d}_{q,u} = \mathbf{q} - \mathbf{q}_{\text{limit up}} \,\,\, \text{and} \,\,\, \mathbf{d}_{q,l} = \mathbf{q} - \mathbf{q}_{\text{limit low}} \, .
\end{equation}
\paragraph{Repelling Force Determination}
First, the acting points $\mathbf{U}_{\text{act},i}$ and $\mathbf{U}_{\text{act},j}$ and acting directions $\mathbf{W}_{\text{act},i}$ and $\mathbf{W}_{\text{act},j}$ of the repelling force are derived by
\begin{IEEEeqnarray}{rclrcl}
\mathbf{U}_{\text{act},i} & = & \mathbf{U}_{i,j}+r_i\tfrac{\mathbf{W}_{i,j}}{\| \mathbf{W}_{i,j}\|} \, ,&\mathbf{W}_{\text{act},i}&=&\mathbf{W}_{i,j}\, ,\\
\mathbf{U}_{\text{act},j} & = & \mathbf{U}_{i,j}+(r_i+d_{i,j})\tfrac{\mathbf{W}_{i,j}}{\| \mathbf{W}_{i,j}\|} \, ,&\,\mathbf{W}_{\text{act},j}&=&-\mathbf{W}_{i,j}\, .
\end{IEEEeqnarray}
They change smoothly over time due to their real-time calculation and the continuous object descriptions using PS. The force magnitude is based on the twice differentiable potential field \cite{Dietrich2011}
\begin{equation}
    E_{\text{rep},i,j} =
    \begin{cases} 
        -\frac{F_\text{max}}{3d_\text{th}^2}(d_{i,j}-d_\text{th})^3, & \text{if }d_{i,j}<d_\text{th}, \\
        0,  &\text{otherwise}.
    \end{cases} \,
\end{equation}
Its first derivative provides the repelling force 
\begin{equation}
\label{eq:f_rep}
    F_{\text{rep},i,j} =
    \begin{cases} 
        -\frac{F_\text{max}}{d_\text{th}^2}(d_{i,j}-d_\text{th})^2, & \text{if }d_{i,j}<d_\text{th}, \\
        0,  &\text{otherwise},
    \end{cases} \,
\end{equation}
while its second derivative defines the stiffness
\begin{equation}
    K_{\text{rep},i,j} =
    \begin{cases} 
        -\frac{2F_\text{max}}{d_{th}^2}(d_{i,j}-d_\text{th}), & \text{if }d_{i,j}<d_\text{th}, \\
        0,  &\text{otherwise},
    \end{cases} \,
\end{equation}
of the virtual spring.
The design variables $F_\text{max}$ and $d_\text{th}$ correspond to the maximum force applied to each primitive and the threshold distance at which the repelling force starts to act. 
The latter is designed to depend on $\dot{d}_{i,j}$ of the potential field entry time step and is chosen higher for higher velocities and vice versa. This allows closer distances for lower velocities, which is useful when, e.g., moving through tight openings. 
Please note that environment-robot distance pairs result in one repelling force while robot-robot ones result in two, as repelling forces can only be applied to actuatable parts.
The repelling force for the joint limit avoidance is determined using the same equations. However, since their distances are already given in joint space, \eqref{eq:f_rep} yields repelling torques instead of Cartesian forces.
\paragraph{Damping Design}
The robot mass $m_{d,i}$ acting along $\mathbf{W}_{\text{act}}$ requires damping for performance and energy dissipation reasons. We calculate it online using
\begin{equation}
m_{d,i} = (\mathbf{J}_{d,i}\mathbf{M}^{-1}(\mathbf{q})\mathbf{J}_{d,i}^T)^{-1} \,,
\end{equation}
where $\mathbf{J}_{d,i} \in \mathbb{R}^{1\times N}$, the projection of the translational geometric Jacobian $\mathbf{J}_{x,i} \in \mathbb{R}^{3\times N}$ of $\mathbf{U}_{\text{act},i}$, is derived by 
\begin{equation}
\mathbf{J}_{d,i} = \tfrac{\mathbf{W}_{\text{act}}^T}{\| \mathbf{W}_{\text{act}}\|}\mathbf{J}_{x,i} \, .
\end{equation}
The determination of Jacobians as $\mathbf{J}_{x,i}$ for custom points of a robot is described in \cite{Siciliano.2010}.
Finally, the damping $D$ is obtained according to the double diagonalization approach \cite{Albu-Schaffer2004}, using
\begin{equation}
D = 2m_{d,i}\sqrt{K_{\text{rep},i,j}}\zeta \, ,
\end{equation}
where the design value $\zeta \in [0,1]$ is the damping constant.
\paragraph{Joint Space Transformation}
Given all dampings, we use the velocity $\dot{d} = \mathbf{J}_{d,i}\dot{\mathbf{q}}$ of $\mathbf{U}_{\text{act}}$ projected along $\mathbf{W}_{\text{act}}$ to calculate the damping force $F_D=D\dot{d}$ and subtract this from the corresponding repelling force. The result is transformed into joint torques using the Jacobian transpose and then summed up according to 
\begin{equation}
\boldsymbol{\tau}_\text{rep} = \sum_{i\neq j} \mathbf{J}_{d,i,j}^{T}(F_{\text{rep},i,j} - D_{i,j}\dot{d}_{i,j}) \, ,
\end{equation}
$\forall  i \in  \{0,\dots, N_{\text{PS}_\text{act}} \},  j \in  \{0,\dots, N_{\text{PS}_\text{all}}\}$, where $N_{\text{PS}_\text{act}}$ is the number of actuatable PSs and $N_{\text{PS}_\text{all}}$ is the number of all PSs.
Finally, $\boldsymbol{\tau}_\text{rep}$ is fed into the control loop, as shown in Fig. \ref{fig:control_loop}.
\section{EXPERIMENTAL VALIDATION}


\subsection{Experimental Setup}
We carried out our experiments with Intelligent Robotic Lab Assistants \cite{Knobbe2022} where the FE robot and its y-z gantry system are controlled in one common \SI{1}{\kilo\hertz} control loop. 
In the experiments showcasing our pipeline from the twin to the real world, we use a preconstructed virtual environment representation. Update procedures to this are prospect of our future work.
The design parameters used are given in Table \ref{tab:DesignValues}.
\begin{table}[tpb]
    \centering
    \caption{Design values used in the experiments}
    \vspace{-0.2cm}
    \begin{tabular}{lccc}
        \toprule
        Parameter                   & Symbol    & Value & Unit          \\ 
        \midrule
        Max repelling force     & $F_\text{max}$ & 30    & \si{\newton}  \\
        Distance threshold          & $d_\text{th}$  & 0.06  & \si{\meter}   \\
        Damping constant            & $\zeta$   & 0.2   & $\frac{\sqrt{\text{N}}\text{\,s}}{\sqrt{\text{m}}\text{\,kg}}$   \\
        Primitive costs            & $[c_{\text{abox}};c_{\text{obox}};c_{\text{sph}};c_{\text{cap}}]$   & [2;2;0.9;1]   & $1$   \\
        \bottomrule
    \end{tabular}
    \vspace{-0.6cm}
    \label{tab:DesignValues}
\end{table}
We carried out five experiments where we investigated our proposed pipeline, the self-collision avoidance behavior, the effect of the damping constant $\zeta$ in combination with the environment collision avoidance behavior, the joint limit avoidance behavior, and lastly, we show how the intelligent twin decisions can be deployed.
\subsection{Experimental Results}
\begin{figure}[tpb]
    \centering
\begingroup%
  \makeatletter%
  \providecommand\color[2][]{%
    \errmessage{(Inkscape) Color is used for the text in Inkscape, but the package 'color.sty' is not loaded}%
    \renewcommand\color[2][]{}%
  }%
  \providecommand\transparent[1]{%
    \errmessage{(Inkscape) Transparency is used (non-zero) for the text in Inkscape, but the package 'transparent.sty' is not loaded}%
    \renewcommand\transparent[1]{}%
  }%
  \providecommand\rotatebox[2]{#2}%
  \newcommand*\fsize{\dimexpr\f@size pt\relax}%
  \newcommand*\lineheight[1]{\fontsize{\fsize}{#1\fsize}\selectfont}%
  \ifx\svgwidth\undefined%
    \setlength{\unitlength}{244.7518158bp}%
    \ifx\svgscale\undefined%
      \relax%
    \else%
      \setlength{\unitlength}{\unitlength * \real{\svgscale}}%
    \fi%
  \else%
    \setlength{\unitlength}{\svgwidth}%
  \fi%
  \global\let\svgwidth\undefined%
  \global\let\svgscale\undefined%
  \makeatother%
  \begin{picture}(1,0.61566957)%
    \lineheight{1}%
    \setlength\tabcolsep{0pt}%
    \put(0,0){\includegraphics[width=\unitlength,page=1]{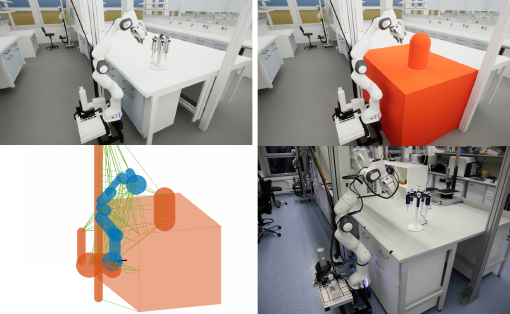}}%
    \put(0.46264703,0.34893971){\makebox(0,0)[t]{\lineheight{1.25}\smash{\begin{tabular}[t]{c}(a)\end{tabular}}}}%
    \put(0.46264684,0.01873548){\color[rgb]{0,0,0}\makebox(0,0)[t]{\lineheight{1.25}\smash{\begin{tabular}[t]{c}(c)\end{tabular}}}}%
    \put(0.96430423,0.34894064){\color[rgb]{0,0,0}\makebox(0,0)[t]{\lineheight{1.25}\smash{\begin{tabular}[t]{c}(b)\end{tabular}}}}%
    \put(0.96430497,0.01873473){\color[rgb]{0,0,0}\makebox(0,0)[t]{\lineheight{1.25}\smash{\begin{tabular}[t]{c}(d)\end{tabular}}}}%
  \end{picture}%
\endgroup%

    \caption{Visualization of our collision avoidance pipeline.}
    \label{fig:Pipeline}
    \vspace{-0.6cm}
\end{figure}
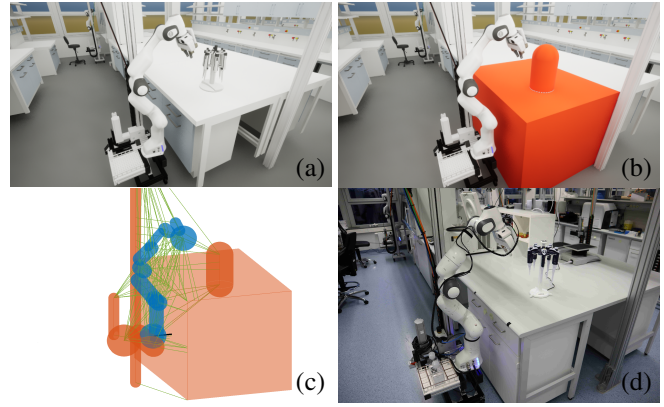

Fig. \ref{fig:Pipeline} visualizes the four steps of our pipeline. Based on the initial situation in \ref*{fig:Pipeline}a, the RAL auto-generates environment PSs of the reachable meshes, resulting in \ref*{fig:Pipeline}b.
The real-time collision avoidance algorithm, visualized in \ref*{fig:Pipeline}c, complements the environment PSs with the robot's ones and determines minimal distances plotted in green if $d_{i,j}\leq d_\text{th}$ and in black if $d_{i,j}>d_\text{th}$. If necessary, it then applies path-correcting torques to the real-world robot shown in \ref*{fig:Pipeline}d.

\begin{figure*}[tpb]
    \centering
    \begin{subfigure}{0.24\textwidth}
        \centering
        \small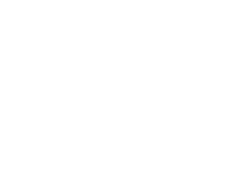
        \subcaption{Absolute end effector velocity}
        \label{subfig:Ex1:v_EE}
    \end{subfigure}
    \begin{subfigure}{0.24\textwidth}
        \centering
        \small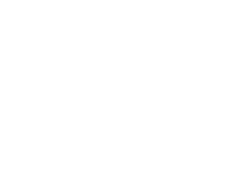
        \subcaption{Surface distance}
        \label{subfig:Ex1:d_EE}
    \end{subfigure}
    \begin{subfigure}{0.24\textwidth}
        \centering
        \small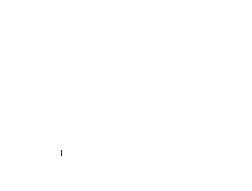
        \subcaption{Repelling and damping forces}
        \label{subfig:Ex1:f_EE}
    \end{subfigure}
    \begin{subfigure}{0.24\textwidth}
        \centering
        \small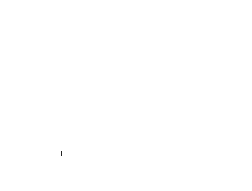
        \subcaption{Resulting torques}
        \label{subfig:Ex1:t_EE}
    \end{subfigure}
    \vspace{-0.2cm}
    \caption{Results of the self-collision experiment between end effector and joint 2 of the FE robot. (a) plots the end effector velocity and (b) the distance between both primitives. The vertical lines indicate the time of entering and leaving of the APF. The horizontal line is $d_\text{th}$. (c) shows the repelling and damping forces for both objects and (d) plots the resulting torques.}
    \label{fig:Ex1}
    \vspace{-0.55cm}
\end{figure*}
The results of the self-collision experiment are presented in Fig. \ref{fig:Ex1}. Depending on the surface distance, repelling and damping forces are applied to both collision parts but in opposite directions. The damping force supports the potential field before reaching the minimal surface distance and decelerates the robot while leaving. This can be seen in Fig. \ref{subfig:Ex1:f_EE} and in Fig. \ref{subfig:Ex1:v_EE}, where the exit velocity is lower than that of the entering time step. The torques in Fig. \ref{subfig:Ex1:t_EE} caused by $F_{\text{rep},EE}$ are different from those caused by $F_{\text{rep},2}$ although the forces have similarities. The different Jacobian matrices explain this. Link 2 can only be moved by the first joint, while the end-effector pose depends on all joints. 

\begin{figure}[tpb]
    \centering
    \begin{subfigure}{\columnwidth}
        \centering
        \small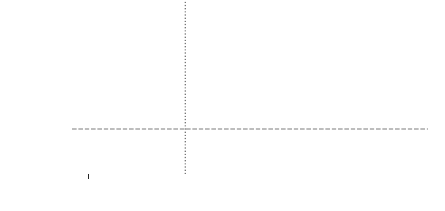
        \subcaption{Surface distance}
        \label{subfig:Ex2:d_EE}
    \end{subfigure}
    \begin{subfigure}{\columnwidth}
        \centering
        \small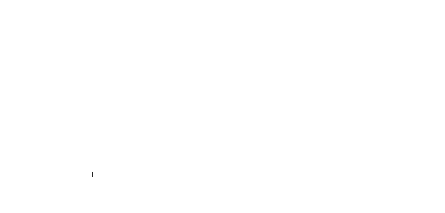
        \subcaption{Torques of joint 3}
        \label{subfig:Ex2:t_EE}
    \end{subfigure}
    \vspace{-0.4cm}
    \caption{Results of the damping experiment. The collision objects are the end-effector and an incubator wall. (a) plots the surface distance and (b) the generated torque on joint 3.}
    \label{fig:Ex2}
    \vspace{-0.6cm}
\end{figure}

In the damping experiment, damping constants $\zeta = \{0,\,0.05,\,0.1,\,0.2,\,0.5\}$ were tested. Fig. \ref{subfig:Ex2:d_EE} shows the surface distance between the end-effector and incubator wall for high-velocity collisions. The potential field entry velocity of the end-effector is the same for all damping constants, and we synchronized the time of entry. Fig. \ref{subfig:Ex2:t_EE} plots the generated torque on joint 3, chosen as it applies the highest torque in this constellation. The experiment was performed with the same joint constellation for each damping constant, making the torques comparable. Although the damping forces increase with the damping constant, $\zeta= 0.5$ does not provide a higher damping effect than $\zeta= 0.2$. This is because of the robot limits for $\dot{\boldsymbol{\tau}}$. The high velocities of the experiment, in combination with the high damping constant, create discontinuities the robot can not follow. Also noticeable is that the undamped system collides with the incubator wall and that a small damping of $\zeta= 0.05$ already prevents this collision. $\zeta = 0.2$ provides the best results and is used in the other experiments.

\begin{figure}[tpb]
    \centering
    \begin{subfigure}{0.49\columnwidth}
        \centering
        \small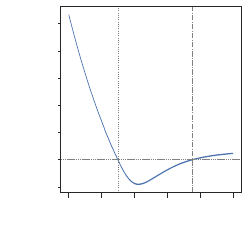
        \subcaption{Angular distance joint 4}
        \label{subfig:Ex3:d_EE}
    \end{subfigure}
    \begin{subfigure}{0.49\columnwidth}
        \centering
        \small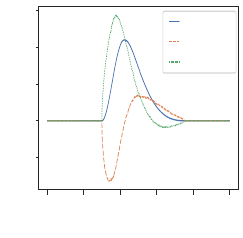
        \subcaption{Torques of joint 4}
        \label{subfig:Ex3:t_EE}
    \end{subfigure}
    \vspace{-0.4cm}
    \caption{Results of the joint limit avoidance experiment. (a) plots the distance between $q_4$ and its lower limit, and (b) shows the torques generated by this collision.}
    \label{fig:Ex3}
    \vspace{-0.65cm}
\end{figure}

During the joint limit avoidance experiment, the angular distance between joint 4 $q_4$ and its lower limit was recorded together with the repelling torque $\tau_{4,\text{rep}}$, the damping torque $\tau_{4,d}$ and their acting difference $\tau_{4}$. The results are shown in Fig. \ref{fig:Ex3}. The generated torques are smooth, and the damping dissipates the kinetic energy excess so that $q_4$ leaves the potential field without being thrown out.

\begin{figure}[tpb]
    \vspace{-0.1cm}
    \centering
    \begin{subfigure}{\columnwidth}
        \centering
        \small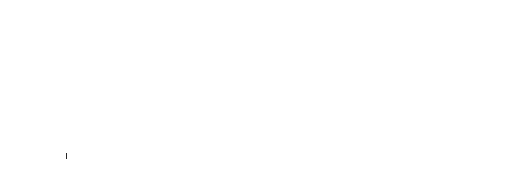
        \subcaption{Distance between the z-axis and end-effector}
        \label{subfig:Ex4:d_EE}
    \end{subfigure}
    \begin{subfigure}{\columnwidth}
        \centering
        \small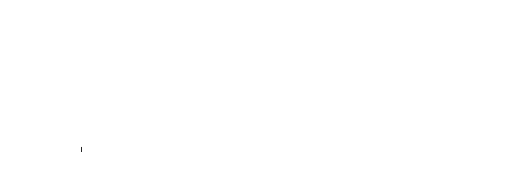
        \subcaption{Repelling forces and velocities}
        \label{subfig:Ex4:t_EE}
    \end{subfigure}
    \vspace{-0.45cm}
    \caption{Results of the evasive base motion experiment. (a) plots the distance between the z-axis and end-effector along with the normalized y-entry of the distance vector. (b) plots the repelling forces.}
    \label{fig:Ex4}
    \vspace{-0.22cm}
\end{figure}

We demonstrate the benefit of deploying intelligent decisions in a task where the gantry system is not required to remain in its position.
So the RAL sets the gantry system to be actuatable, and the control changes its behavior accordingly. Fig. \ref{fig:Ex4} plots the resulting gantry system evasive motions. 
The upper graph plots the distance $d_{EE,Z}$ between the surfaces of the end-effector (EE) and z-axis (Z) along with $w_{EE,Z_{0,y}}$, the y-entry of the normalization to length 1 of $\mathbf{W}_{EE,Z}$. As the z-axis can only be moved by the y-axis motor, collision-avoiding motions of the z-axis are only possible in y-direction. Therefore, the repelling force $F_{\text{rep},Z} = -20w_{EE,Z_{0,y}}F_{\text{rep},EE}$ is scaled by the plotted $w_{EE,Z_{0,y}}$. Factor 20 was introduced to compensate for the higher mass of the z-axis compared to the robot links.
During the experiment, the end-effector was moved from the left side of the z-axis to the right and back, as it can be recognized by the change of sign of $w_{EE,Z_{0,y}}$. 
The lower diagram plots the repelling forces $F_{\text{rep},Z}$ and $F_{\text{rep},EE}$ and the base velocity $v_{O}$.
The base velocity $v_{O}$ turned out to be a scaled version of $F_{\text{rep},Z}$, showing that the evasive base motion works. The change in sign of $v_{O}$ and $F_{\text{rep},Z}$ implies its functionality in both directions.

\subsection{Discussion}
The experiments showed the functionality of our proposed collision avoidance pipeline. The RAL algorithm is convergent, its PSs are tight, and our APF implementation avoids collisions of all types from Fig. \ref{fig:collision_taxonomy}.

One weakness of APFs, however, is the possibility of local minima where the robot can get stuck. To overcome this, we suggest improving the RAL algorithm by taking advantage of the twin's knowledge about the process, robot, and environment. For example, when the robot is not intended to move into open or u-shaped objects, the RAL can represent them as a closed box or sphere. Embedding such intelligence in control is difficult due to real-time constraints. With our proposed pipeline, we can, however, intelligently adjust the collision avoidance behavior by making decisions outside control in our knowledge base where we have the required knowledge and computation power.

A strength of APFs, in turn, is their real-time capability. Our whole real-time algorithm, including state update and distance and torque calculations, takes \SI{7.45}{\micro\second} on average in the scenario of Fig. \ref{fig:Pipeline}.
However, this time can merely be used to get an idea of the order of its magnitude since it depends on the types and numbers of PSs and distance pairs, as well as the number of distances below $d_\text{th}$, and the frame to which the PSs with $d_{i,j}<d_\text{th}$ are linked. In Fig. \ref{fig:Pipeline}, we had 4 spheres, 13 capsules, 3 planes, 126 distance pairs of all types, and 1 distance with $d_{i,j}<d_\text{th}$ between base and world frame.
When adding more PSs $n \in \mathbb{N}$ to the algorithm, its time complexity grows most in the parts where computations are performed for each of the $\frac{(n-1)n}{2}-m \propto \mathcal{O}(n^2)$ distance pairs, where $m \in \mathbb{N}$ is the number of distance pairs that cannot collide due to the structure, e.g., PSs that are linked to the same frame. 
However, after approximating the 9-DOFs robot, i.e., all movable parts, with 4 spheres and 11 capsules, the only PSs the RAL keeps adding and removing are those from tools, grasped objects, and the environment.
While the comparatively few tool or grasped object PSs that move with the robot affect the time complexity according to the above rate, the vast majority of added PSs represent the fixed environment that cannot collide with itself, thus requiring only environment-robot distances, and therefore causing a linear time complexity $\mathcal{O}(n)$.
Consequently, with the exception of the few tool and grasped object PSs, we can say that our algorithm has a linear time complexity. 

\begin{figure}[tpb]
    \centering
    \begin{subfigure}{0.49\columnwidth}
        \def\svgwidth{\columnwidth}
        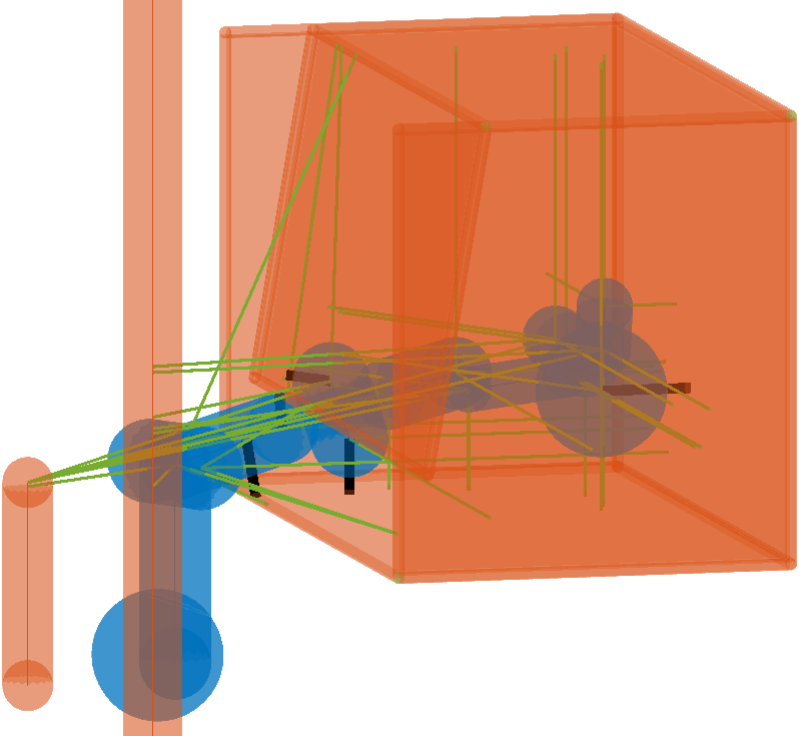
    \end{subfigure}
    \begin{subfigure}{0.49\columnwidth}
        \def\svgwidth{\columnwidth}
        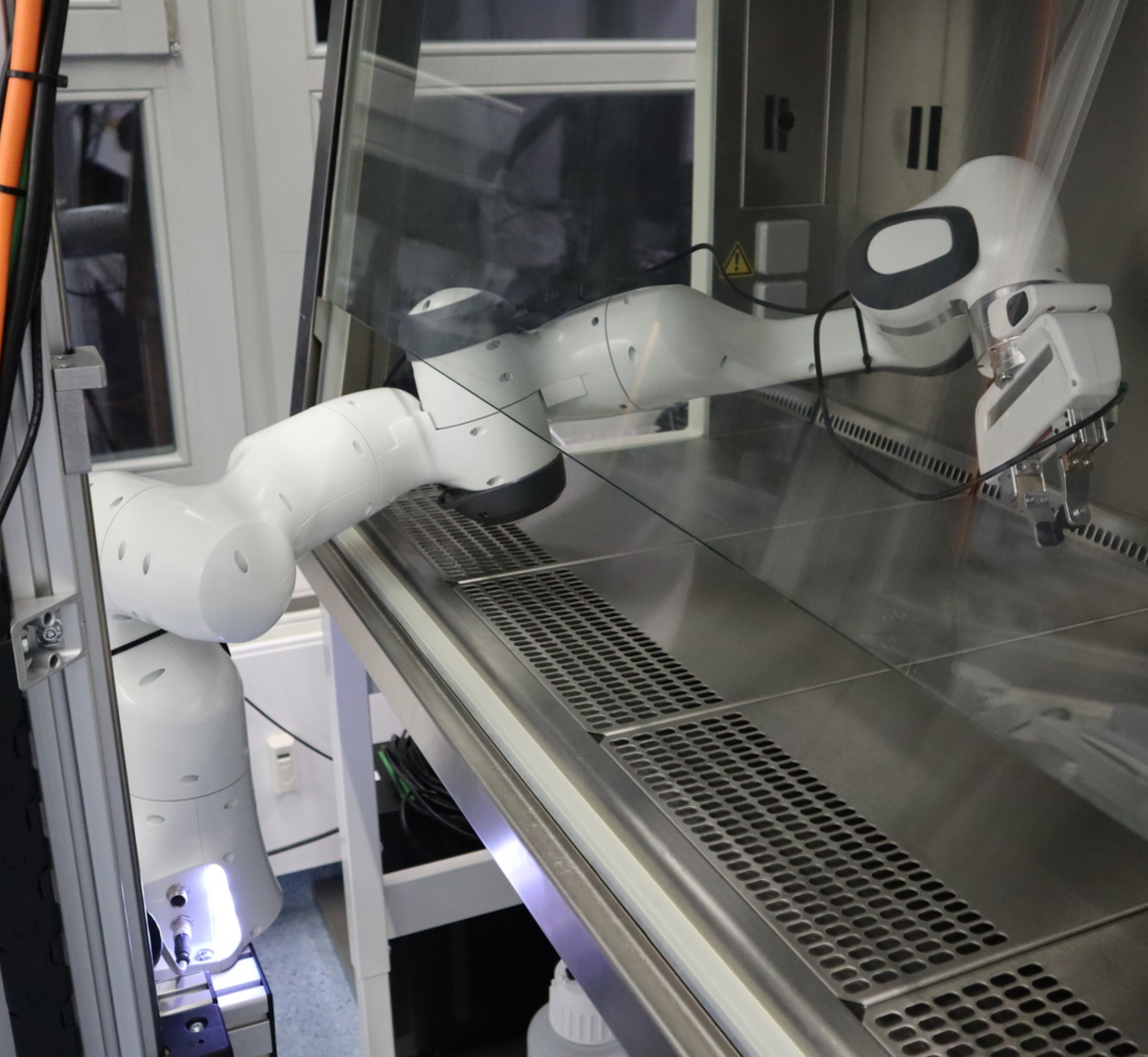
    \end{subfigure}
    \caption{The robot working in confined space.}
    \label{fig:DigitalLab}
    \vspace{-0.65cm}
\end{figure}

We discovered many possible applications for our system. An example is given in Fig. \ref{fig:DigitalLab}, where the robot operates safely in a confined space. While working in the biosafety cabinet, the robot link in the opening is perfectly leveled in the middle, and the rest of the arm avoids collisions inside. The whole system behaves stable and robustly.




\section{CONCLUSIONS AND FUTURE WORKS}
Our proposed pipeline achieves environment-aware, high-performance, whole-body collision avoidance in \SI{1}{\kilo\hertz} real-time. The deployment of task-grounded intelligent decisions was enabled by connecting control to a digital twin instead of a stand-alone data processing algorithm. This keeps the ability of intentional manipulation and adds the possibility of adapting the collision avoidance behavior from outside real-time control, where more background knowledge is available.
In the future, we plan to improve the RAL algorithm to make collision avoidance even smarter. First, we can further reduce the number of considered PSs by providing only those reachable within the next second, and second, we aim to avoid the occurrence of local minima by choosing the PS type task-grounded. 
Moreover, supplementing the twin with an update procedure, allowing for the avoidance of dynamic obstacles, is a prospect of future work.

\bibliographystyle{IEEEtran} 
\bibliography{References/literature.bib}

\end{document}